\documentclass[final,5p,times]{elsarticle}

\usepackage{graphicx}
\usepackage{amsmath}
\usepackage{amssymb}
\usepackage{hyperref}
\usepackage{lineno}
\usepackage{subcaption}
\usepackage{multirow}
\usepackage{dsfont}
\usepackage{float}  
\usepackage[superscript]{cite}

\usepackage{booktabs} 
\modulolinenumbers[5]

\newcommand{\suppMethodsSec}[1]{Supplementary Section #1 }
\newcommand{\suppMethodsFig}[1]{Supplementary Figure #1}
\newcommand{\suppMethodsTab}[1]{Supplementary Table #1 }

\newcommand{\appTabCohortStats}{\suppMethodsTab{1}} 
\newcommand{\appCnnArchitecture}{\suppMethodsSec{2}} 
\newcommand{\appCnnSchematic}{\suppMethodsFig{1}} 
\newcommand{\appSegmentationArchitecture}{\suppMethodsSec{3}} 
\newcommand{\appSegmentationSchematic}{\suppMethodsFig{2}} 
\newcommand{\appModelSelection}{\suppMethodsSec{4}} 

\biboptions{sort&compress,super}

\journal{The Lancet Digital Health}

\begin{document}

\begin{frontmatter}
\title{Detection and Localization of Subdural Hematoma Using Deep Learning on Computed Tomography}

\author{
Vasiliki Stoumpou, BSc$^{1}$; 
Rohan Kumar, BSc$^{1,2}$;
Bernard Burman, BSc$^{3}$;\\
Diego Ojeda, MD$^{4}$; 
Tapan Mehta, MD, MPH$^{4,5}$; 
Dimitris Bertsimas, PhD$^{1,6}$ \\[1em]
\small $^{1}$Operations Research Center, Massachusetts Institute of Technology, Cambridge, MA, USA \\
\small $^{2}$Boston University, Boston, MA, USA \\
\small $^{3}$Massachusetts Institute of Technology, Cambridge, MA, USA \\
\small $^{4}$University of Connecticut School of Medicine, Farmington, CT, USA \\
\small $^{5}$Hartford HealthCare, Hartford, CT, USA \\
\small $^{6}$Sloan School of Management, Massachusetts Institute of Technology, Cambridge, MA, USA \\
\small *Corresponding author: Dimitris Bertsimas \\ Sloan School of Management, Massachusetts Institute of Technology, Cambridge MA, 02142 \\ \texttt{dbertsim@mit.edu}
}




\begin{abstract}





\noindent
Background.
Subdural hematoma (SDH) is a common neurosurgical emergency, with increasing incidence in aging populations. Rapid and accurate identification is essential to guide timely intervention, yet existing automated tools focus primarily on detection and provide limited interpretability or spatial localization. There remains a need for transparent, high-performing systems that integrate multimodal clinical and imaging information to support real-time decision-making.

\noindent
Methods.
We developed a multimodal deep-learning framework that integrates structured clinical variables, a 3D convolutional neural network trained on CT volumes, and a transformer-enhanced 2D segmentation model for SDH detection and localization. Using 25,315 head CT studies from Hartford HealthCare (2015–2024), of which 3,774 (14·9\%) contained clinician-confirmed SDH, tabular models were trained on demographics, comorbidities, medications, and laboratory results; imaging models were trained to detect SDH and generate voxel-level probability maps. A greedy ensemble strategy combined complementary predictors. The primary outcome was area under the receiver operating characteristic curve (AUC) for SDH detection; secondary outcomes included Dice coefficient and qualitative interpretability of localization outputs.

\noindent
Findings.
Clinical variables alone provided modest discriminatory power (AUC 0·75). Convolutional models trained on CT volumes and segmentation-derived maps achieved substantially higher accuracy (AUCs 0·922 and 0·926). The multimodal ensemble integrating all components achieved the best overall performance (AUC 0·9407; 95\% CI, 0·930–0·951) and produced anatomically meaningful localization maps consistent with known SDH patterns.

\noindent
Interpretation.
This multimodal, interpretable framework provides rapid and accurate SDH detection and localization, achieving high detection performance and offering transparent, anatomically grounded outputs. Integration into radiology workflows could streamline triage, reduce time to intervention, and improve consistency in SDH management.

\noindent
Funding.
The work was supported by institutional resources from the Massachusetts Institute of Technology and Hartford HealthCare.

\end{abstract}


\end{frontmatter}

\newpage
\section{Introduction}
\noindent
A subdural hematoma (SDH) is an intracranial bleed occurring in the subdural space, between the skull and the brain, and can cause mass effect and neurological deterioration. SDHs are broadly classified by acuity into acute, subacute, and chronic types; acute SDH often results from traumatic injury, whereas chronic SDH may evolve slowly over weeks or months. Risk factors include increasing age, head trauma (even mild), anticoagulant or antiplatelet use, coagulopathies, and alcohol use. 

The incidence of SDH and chronic SDH is increasing; estimates suggest that over 90,000 hospitalizations per year in the U.S. are attributed to SDH \cite{frontera2011national}, and approximately 60,000 Americans annually may develop chronic SDH alone \cite{balser2015actual}. Chronic SDH incidence rates in the literature range from around 1·7 to 20·6 per 100,000 persons per year, with much higher rates in the elderly population \cite{rauhala2019incidence, chen2024advances}. Mortality for acute traumatic SDH remains significant; acute SDH mortality remains high (14–50\% across cohorts) \cite{ryan2012acute, fluss2023deadliness}, whereas chronic SDH carries lower mortality but up to 20\% recurrence after surgical evacuation \cite{ramachandran2007chronic, miah2021radiological, oh2010postoperative}.

Diagnostic inaccuracy or delayed intervention can worsen neurological outcomes of SDH. Several studies have demonstrated that delayed surgical intervention is associated with poorer neurological recovery and longer hospitalization durations; for example, patients undergoing surgery within 3 days after admission achieve significantly better postoperative outcomes and shorter stays compared to those treated later \cite{colonna2025impact}, while timely evacuation in traumatic SDH has been historically linked to mortality reductions when performed within hours of injury \cite{tien2011reducing, chen2020impact}. In parallel, recent trials of MMA embolization as adjunctive therapy have shown reduced recurrence rates and reinterventions in chronic SDH patients, highlighting its growing role in SDH management \cite{davies2024adjunctive, salih2022reduced, debs2024middle, moshayedi2020middle}.

Early and accurate detection and localization of SDH is thus a critical first step—not only to guide surgical timing and intervention choice (e.g. burr-hole, craniotomy, or MMA embolization) but to institutionalize consistent triage practices and documentation workflows across imaging and electronic health record systems.

In this context, AI and deep learning have become integral to medical imaging workflows, enabling automated detection, segmentation, and quantification of pathologies across multiple organ systems \cite{pinto2023artificial, khalifa2024ai}. In parallel, the increase in data availability has led to the leverage of multiple modalities for a variety of predictive tasks, with works such as the HAIM framework \cite{soenksen2022integrated} reporting the benefits of multimodal fusion. In the domain of intracranial hemorrhage, many works focus on parenchymal bleeds (ICH) \cite{seyam2022utilization,abed2025artificial, rohren2025post}, but comparatively fewer address extra-axial bleeding such as subdural hematoma (SDH).

Several FDA-cleared AI systems have recently been introduced for automated subdural hematoma (SDH) detection on head CT, most notably \textit{Rapid~SDH} (RapidAI) and \textit{Viz~Subdural} (Viz.ai). These commercial platforms identify suspected SDH and generate volumetric estimates within minutes of image acquisition \cite{colasurdo2022automated}. Rapid SDH is reported by RapidAI to achieve ~92 \% sensitivity for SDH detection, while Viz~Subdural has shown similar diagnostic accuracy and high agreement with clinician measurements of hematoma volume and midline shift \cite{colasurdo2022automated}.

However, commercial tools like Rapid SDH face limitations and lack transparency. Their published performance is often constrained to detection (presence/absence) rather than pixel-level segmentation or volumetric localization. Even when a suspected hematoma region is highlighted, quantitative evaluation of segmentation accuracy is not reported. They tend to focus on SDH above a certain volume threshold (e.g. $>1 mL$), potentially missing small or subtle bleeds. Moreover, many commercial systems do not expose internal model details, training datasets, or strategies for combining multimodal clinical data. 

Colasurdo et al.\ \cite{colasurdo2022automated} validated the Viz.ai SDH platform on a small external dataset. While demonstrating strong concordance with manual measurements, their study was limited by a small, single-center retrospective cohort of a few hundred patients and excluded technically inadequate or severely deformed cases with large midline shifts.
\begin{center}
\fbox{
\begin{minipage}{0.46\textwidth}
\section*{Research in context}

\noindent\textbf{Evidence before this study}

Existing commercial tools for subdural hematoma (SDH) analysis focus primarily on detection and typically emphasize larger-volume bleeds, with limited transparency regarding model behavior, training data, or performance for small or subtle hematomas. Prior academic work on intracranial hemorrhage detection has shown promising results using deep-learning models applied directly to head CT, but these approaches generally rely on imaging alone and are often evaluated on relatively small sample sizes. As a result, little evidence exists on multimodal pipelines that integrate clinical features with both detection and localization to support real-time SDH assessment, while also having been evaluated on thousands of studies.

\noindent\textbf{Added value of this study}

This study introduces a multimodal deep-learning system that integrates structured clinical data, 3D CT-based classification, and transformer-enhanced 2D segmentation to detect and localize SDH. By combining complementary information from tabular data, volumetric image representations, and pixel-level segmentation maps, the framework provides high diagnostic accuracy while producing clinically interpretable localization outputs. To our knowledge, this is the first SDH pipeline to unify tabular clinical features with two distinct imaging-based models and to evaluate such an approach at the scale of more than 25{,}000 CT studies.

\noindent\textbf{Implications of all the available evidence}

Taken together, existing work suggests substantial room for improvement in automated SDH diagnosis, particularly regarding interpretability and multimodal fusion. Our findings show that integrating clinical and imaging modalities can yield accurate, transparent predictions with detailed spatial maps that may support emergency triage and radiology workflows. These results motivate future prospective validation and exploration of real-time deployment in clinical settings.
\end{minipage}
}
\end{center}

In the research front, Kaya et al.\ \cite{kaya2023revolutionary} proposed a two-tiered system combining Mask R-CNN segmentation and SVM classification for acute SDH detection, achieving high slice-level accuracy, but restricted to 2D images and acute presentations only.
For segmentation, Petrov et al.\ \cite{petrov2024ai} and Farzaneh et al.\ \cite{Farzaneh2020} used U-Net–based architectures on limited datasets (less than 110 scans); the former only focuses on chronic cases, while the latter relies on extensive preprocessing or handcrafted features. Collectively, these works underscore the promise of deep learning for SDH detection and localization but remain constrained by single-modality designs, small sample sizes, and potentially limited generalizability. 

Beyond SDH-specific studies, broader work in intracranial hemorrhage segmentation has underscored the value of deep learning for spatially precise localization. Zarei et al. \cite{zarei2024deep} demonstrated accurate ICH segmentation on noncontrast CT, and recent multitask models, such as SwinTransformer–Swin-Unet hybrids, have shown improved robustness across imaging domains \cite{hirata2025brain}. Meanwhile, general-purpose vision–language models like ChatGPT-4V achieve only modest performance (around 60\% accuracy for SDH detection) even with guided prompts \cite{bayar2025artificial}. These findings highlight that domain-specific, clinically trained architectures remain essential for reliable radiologic interpretation.

\begin{figure*}[t]
    \centering
    \includegraphics[width=\textwidth]{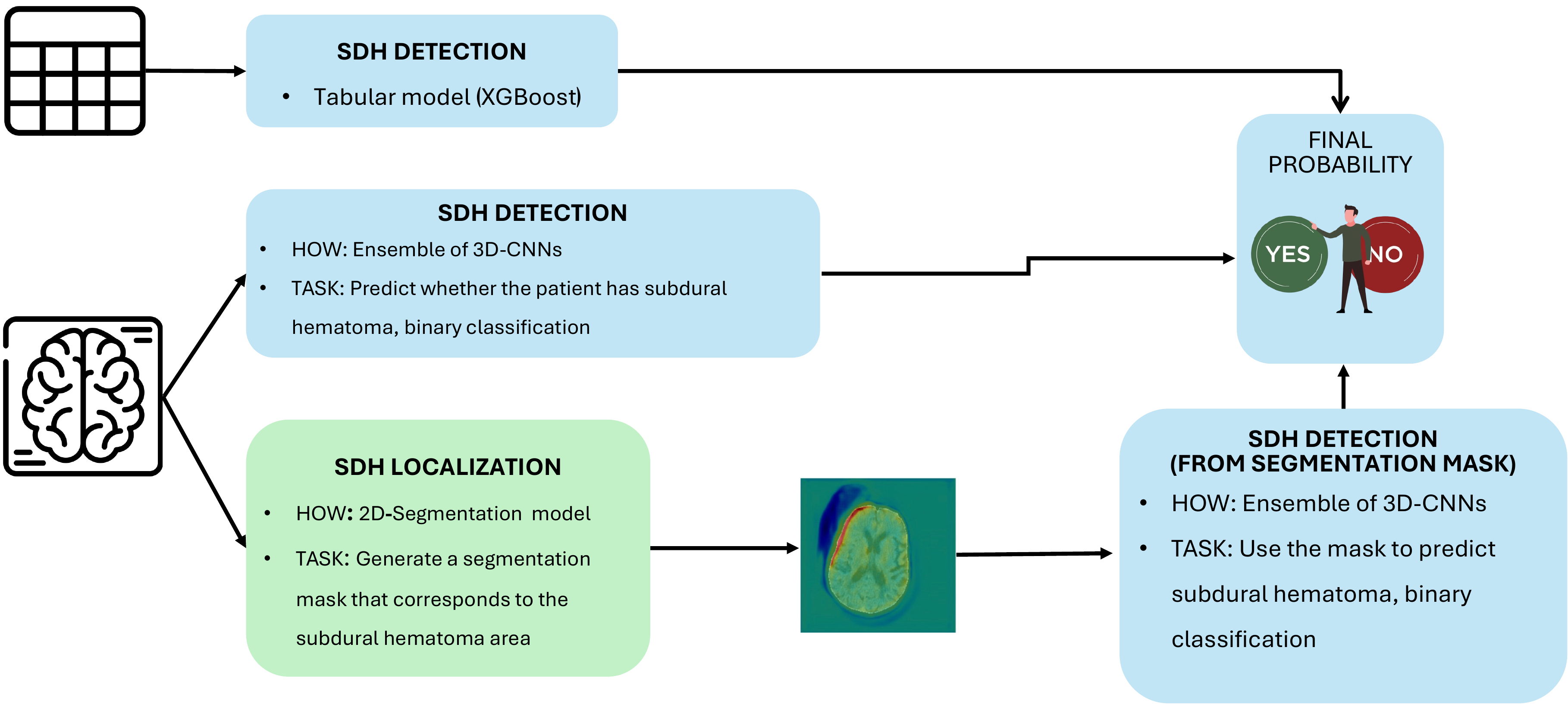}
    \caption{Overview of the proposed multimodal pipeline for subdural hematoma (SDH) detection and localization. 
    The system integrates three complementary components: a tabular XGBoost model using clinical variables, 
    an ensemble of 3D convolutional neural networks (CNNs) applied to CT scans, 
    and a 2D Swin-CNN hybrid segmentation model for hematoma localization. 
    Probabilities from each component are combined into a final ensemble output 
    indicating the likelihood of SDH presence.}
    \label{fig:pipeline_overview}
\end{figure*}

Building on these findings, we propose a novel multimodal pipeline that integrates routinely available clinical information with head CT imaging from more than 25,000 studies. The framework combines volumetric classification with transformer-based segmentation to detect and localize subdural hematomas with high accuracy while providing interpretable SDH probability maps of the brain that are crucial for clinical decision-making. This approach addresses the lack of transparency and the limited detection of smaller hematoma regions which characterize existing solutions, while bridging the performance - interpretability gap often seen in fully black-box systems. By enabling early and explainable detection without reliance on manual measurements, this approach has the potential to reduce delays in treatment, improve the consistency of triage practices, and ultimately enhance neurological outcomes for patients with subdural hematoma. A full schematic of our pipeline is presented in Figure \ref{fig:pipeline_overview}.

\section{Methods}
\noindent
This Section presents the methodology we have developed for performing detection and localization of subdural hematomas. Section \ref{subsec:data} discusses the cohort creation and data description, whereas Section \ref{subsec:detection} presents the model architectures used for detection and localization of the hematomas.

\subsection{Data} \label{subsec:data}
\noindent
We utilized data from Hartford Healthcare, one of the largest hospital systems in Connecticut, comprising information from 15,183 adult patients between 2016 and 2023. Each observation corresponds to one non-contrast head computed tomography (CT) study, while individual patients may contribute multiple studies.

Patients younger than 18 years at the time of imaging, those with a documented history of brain tumor, and those who had undergone craniectomy for reasons other than subdural hematoma are excluded from the analysis. After applying these criteria, the final dataset includes 25,316 imaging studies, of which 3,774 are associated with a confirmed diagnosis of subdural hematoma based on radiology reports and clinical documentation.

Compared with prior studies on subdural hematoma, this dataset represents one of the largest cohorts reported to date, supporting the potential generalizability of our approach \cite{colasurdo2022automated, kaya2023revolutionary, petrov2024ai, Farzaneh2020}. Adopting a multimodal perspective, we integrate both imaging and tabular clinical information for model development.

Data are partitioned into training (64\%), validation (16\%), and test (20\%) sets on the patient level to prevent data leakage across splits.

\subsubsection{Tabular Data}
\noindent
Tabular features include demographic variables (age, sex) and clinical characteristics reflecting patients’ general medical condition. Specifically, we incorporate information on anticoagulant and antiplatelet use (e.g. heparin sodium, aspirin), relevant laboratory values (activated partial thromboplastin time [aPTT], international normalized ratio [INR], platelet count), and the Charlson Comorbidity Index (CCI) components, which capture long-term mortality risk based on comorbid diagnoses such as cancer and diabetes.

For laboratory values, we include both the most recent measurement prior to imaging and an exponentially weighted average of the preceding two years to approximate the temporal evolution of these variables.

\subsubsection{Image Data}
\noindent
All imaging studies consist of three-dimensional axial CT volumes reconstructed from multiple slices with pixel dimensions $512 \times 512$ across the axial plane. CT scans are converted to Hounsfield Units (HU), and windowed to a soft-tissue range (0–130 HU, centered at 65 HU) to enhance subdural contrast \cite{LEV2002427}. The skull is isolated using intensity thresholding and connected-component analysis, and its inverse defines the intracranial region. Slice-level refinement retains the second-largest component, which corresponds to brain tissue.

Because the number of valid slices vary across studies, we standardize all inputs to 15 slices per study by uniformly downsampling scans with more slices and duplicating central slices for scans with fewer than 15. The choice of 15 slices balances information retention and computational feasibility. 


\subsubsection{Cohort Selection}
\noindent
The final cohort includes both case (patients with subdural hematoma) and control (patients without subdural hematoma) imaging studies, as determined by clinical diagnosis and radiology report review. Inclusion and exclusion criteria are applied as described above to ensure a well-defined population for the classification task. More details on the cohort definition can be found in \appTabCohortStats.

\subsection{Model Architecture} \label{subsec:detection}
\noindent

We developed and evaluated several models to detect the presence of subdural hematoma (SDH) from multimodal data, including tabular clinical information, 3D brain CT scans, and segmentation-derived masks. Model performance was assessed on a held-out test set using the area under the receiver operating characteristic curve (AUC) to account for class imbalance.

\subsubsection{Tabular Model}
\noindent
The first component of our pipeline is an XGBoost classifier trained using only tabular features to predict SDH presence. XGBoost was selected for its computational efficiency and strong performance with structured data \cite{chen2016xgboost}. Hyperparameters were optimized through cross-validation, and the model was evaluated with AUC. This model served as a baseline to assess the predictive value of non-imaging variables.

\subsubsection{Convolutional Neural Network} \label{subsubsec: cnns}
\noindent
The second component of the framework is a three-dimensional convolutional neural network (3D CNN) that classifies CT volumes as containing or not containing a subdural hematoma. By extending conventional 2D models to volumetric space, the network captures both within-slice and between-slice spatial patterns that reflect the true three-dimensional nature of hemorrhage. The architecture employs hierarchical feature extraction with residual skip connections to preserve fine anatomical details while learning broader contextual information. To address randomness inherent in training deep architectures, multiple independently initialized networks were ensembled to enhance robustness. More information about the architecture, as well as a detailed layer-wise schematic are provided in \appCnnArchitecture and \appCnnSchematic, respectively.

\subsubsection{Segmentation Model}
\noindent
The detection of a subdural hematoma, while clinically useful, lacks the spatial localization needed for comprehensive evaluation. Hematoma location, size, and morphology are critical for treatment planning and can be inferred from segmentation models that output probability maps indicating, for each pixel, the likelihood of hemorrhage.

To enable this, we manually annotated 335 head CT studies under physician supervision, generating binary masks at 256×256 resolution. Two random spatial augmentations were applied per image to increase variability, yielding over 12,000 slices for training (80\% used for model fitting). An additional 100 control studies with all-zero masks were included to improve discrimination and reduce false positives.

We developed a two-dimensional hybrid segmentation model integrating convolutional and transformer-based feature extraction. The convolutional path, a U-Net variant with spatial and channel attention, captures fine-grained local features, while the Swin Transformer branch models global context such as midline shift or hemispheric asymmetry. Multi-scale features from both branches are fused during decoding to enhance localization accuracy. Compared to a 3D segmentation alternative, this approach achieved similar performance with lower computational cost. More details on the model architecture and training, as well as a detailed architectural schematic, are provided in \appSegmentationArchitecture and \appSegmentationSchematic, respectively.

\subsubsection{CNN on Segmentation Masks}
\noindent
The probability maps produced by the segmentation model were subsequently used as inputs to a separate 3D CNN (with the same architecture as described in Section \ref{subsubsec: cnns}) trained to predict whether a given scan contained a hematoma. This setup allowed the model to learn discriminative patterns from the segmentation - derived representations.

\subsubsection{Model Ensembling}
\noindent
The final ensemble combined eight models: one tabular XGBoost model, five CNNs trained on CT scans, and four CNNs trained on segmentation masks. Adding more models did not have further benefits on the validation set, and was thus omitted. Final predictions were obtained by averaging predicted probabilities across all models to maximize robustness and generalization.



\section{Results}
\noindent
In this Section, we present the Results of our analysis. Section \ref{subsec:cohort_stats} discusses cohort statistics, and Sections \ref{subsec:classification_perf} and \ref{subsec:segmentation_perf} discuss the performance of our approach in the classification and segmentation tasks. Finally, Section \ref{subsec:ensemble} briefly discusses the construction of the ensemble model. 

\subsection{Cohort statistics} \label{subsec:cohort_stats}
\noindent
\appTabCohortStats summarizes cohort characteristics stratified by the presence of subdural hematoma. Patients with subdural hematoma were older on average and more frequently male, consistent with the known epidemiology of the condition. They also exhibited slightly lower platelet counts and coagulation times, and higher rates of anticoagulant or antiplatelet medication use, including aspirin, warfarin, and heparin (all $p <$ 0·001). Most comorbidities from the Charlson Index showed comparable prevalence between groups, indicating that overall health status was similar apart from coagulation-related factors.

\subsection{Classification Performance} \label{subsec:classification_perf}
\noindent
Table~\ref{tab:classification_results} summarizes the performance of all models on the test set. 
The tabular XGBoost model achieved an AUC of 0·75, confirming that clinical variables alone provide limited, but not trivial, discriminatory power for SDH detection.
In contrast, the convolutional models, both those trained directly on CT volumes (CNNs) and those trained on segmentation-derived probability maps (Seg-CNNs), achieved substantially higher performance, with individual AUCs ranging between 0·90 and 0·93. 
Ensembling improved overall robustness and accuracy: the CNN ensemble reached an AUC of 0·922 (95\% CI: 0·909-0·934), while the Seg-CNN ensemble achieved 0·926 (95\% CI: 0·912--0·938). 
Combining all ten models (tabular, CNN, and Seg-CNN) into a single multimodal ensemble yielded the best performance, with an AUC of 0·9407 (95\% CI: 0·930--0·951), exceeding the best individual model by approximately 2\%.

\begin{table*}[t]
\centering
\caption{Performance of individual and ensemble models on the test set. Accuracy, Sensitivity, Specificity and F1-score were calculated using the optimal threshold as calculated by Youden's J statistic on the validation set.}
\label{tab:classification_results}
\begin{tabular}{lcccccc}
\hline
\textbf{Model} & \textbf{AUC (95\% CI)} & \textbf{Accuracy} & \textbf{Sensitivity} & \textbf{Specificity} & \textbf{F1-score} \\
\hline
Overall Ensemble (all 9) & \textbf{0·9407} [0·930--0·951] & \textbf{0·927} & \textbf{0·815} & \textbf{0·947} & \textbf{0·774} \\
CNNs (4-model ensemble) & 0·922 [0·909--0·934] & 0·917 & 0·781 & 0·941 & 0·742 \\
Seg-CNNs (4-model ensemble) & 0·926 [0·912--0·938] & 0·921 & 0·798 & 0·943 & 0·756 \\
Tabular (XGBoost) & 0·751 [0·732--0·770] & 0·675 & 0·671 & 0·675 & 0·388 \\
\hline
\end{tabular}
\end{table*}

\begin{figure*}[t]
    \centering
    \begin{subfigure}[t]{0.48\textwidth}
        \centering
        \includegraphics[width=\textwidth]{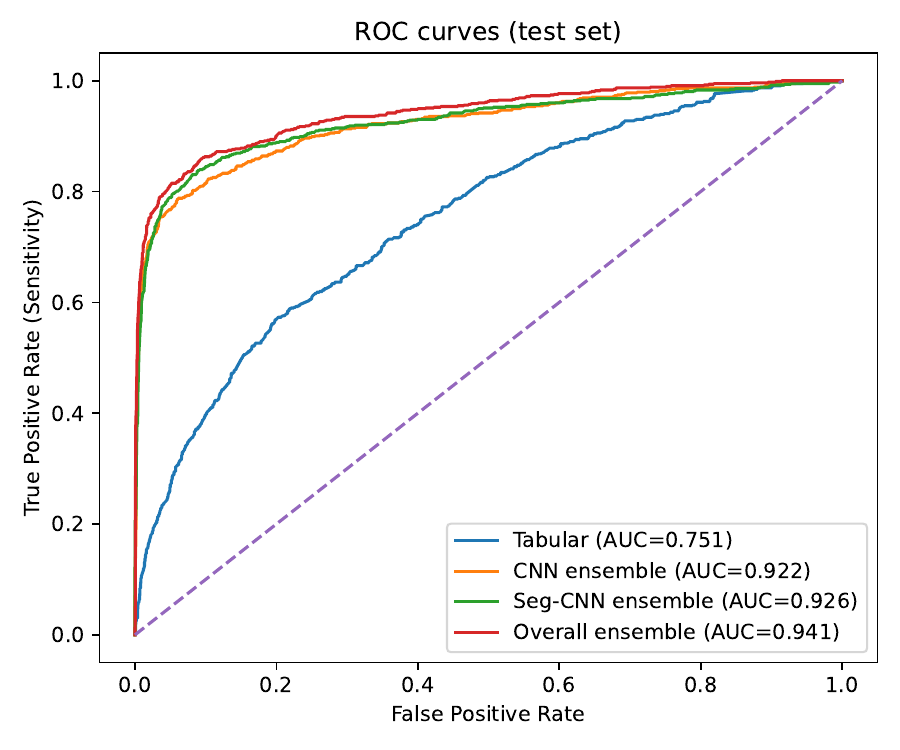}
        \caption{Receiver Operating Characteristic (ROC) curves for the Tabular, CNN, Seg-CNN, and Ensemble models.}
        \label{fig:roc_sub}
    \end{subfigure}
    \hfill
    \begin{subfigure}[t]{0.48\textwidth}
        \centering
        \includegraphics[width=\textwidth]{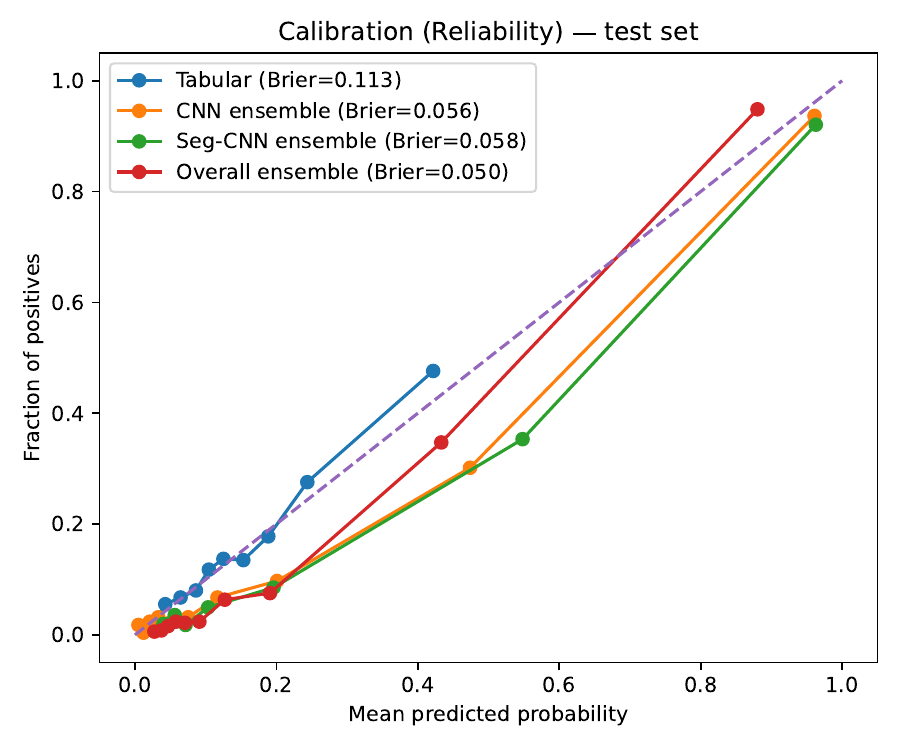}
        \caption{Calibration curves showing agreement between predicted and observed probabilities.}
        \label{fig:calibration_sub}
    \end{subfigure}
    \caption{
        Model evaluation curves on the test set. 
        (a) ROC curves demonstrate strong discrimination (AUCs above 0·90 across all deep learning model groups). 
        (b) Calibration curves show well-aligned predicted and observed probabilities, supporting reliability for clinical use.
    }
    \label{fig:roc_calibration}
\end{figure*}

Figure~\ref{fig:roc_calibration} summarizes model discrimination and calibration on the test set. 
All deep learning models achieved AUCs above 0·90, with the full ensemble reaching 0·94. 
The calibration curves, along with the Brier score of 0·05, demonstrate that predicted probabilities of the ensemble model closely match observed SDH prevalence, indicating good reliability and potential for clinical decision support.

\subsection{Segmentation Performance} \label{subsec:segmentation_perf}
\noindent
Our segmentation model aimed to accurately segment subdural hematomas (SDH) within individual CT slices. Beyond improving detection accuracy, precise segmentation enhances model interpretability and provides a foundation for future work in hematoma quantification and progression tracking.

On the internal validation set, the proposed hybrid 2D segmentation model achieved a Dice score of 0·77, which demonstrates a strong spatial overlap between predicted and ground-truth hematoma regions (1 indicates perfect agreement, 0 indicates no overlap). Despite being trained specifically for SDH, the model generalized effectively to other medical segmentation domains. When retrained and evaluated on the CVC-ClinicDB polyp dataset, it achieved Dice scores between 0·92–0·93, depending on data split. Similarly, on the 2D Brain Tumor Segmentation (Kaggle) dataset \cite{buda2019lggmri}, the model obtained a Dice score of 0·83, outperforming the vanilla U-Net baseline (Dice = 0·76).

\begin{figure*}[t]
    \centering

    \begin{subfigure}{\textwidth}
        \centering
        \includegraphics[width=0.9\textwidth]{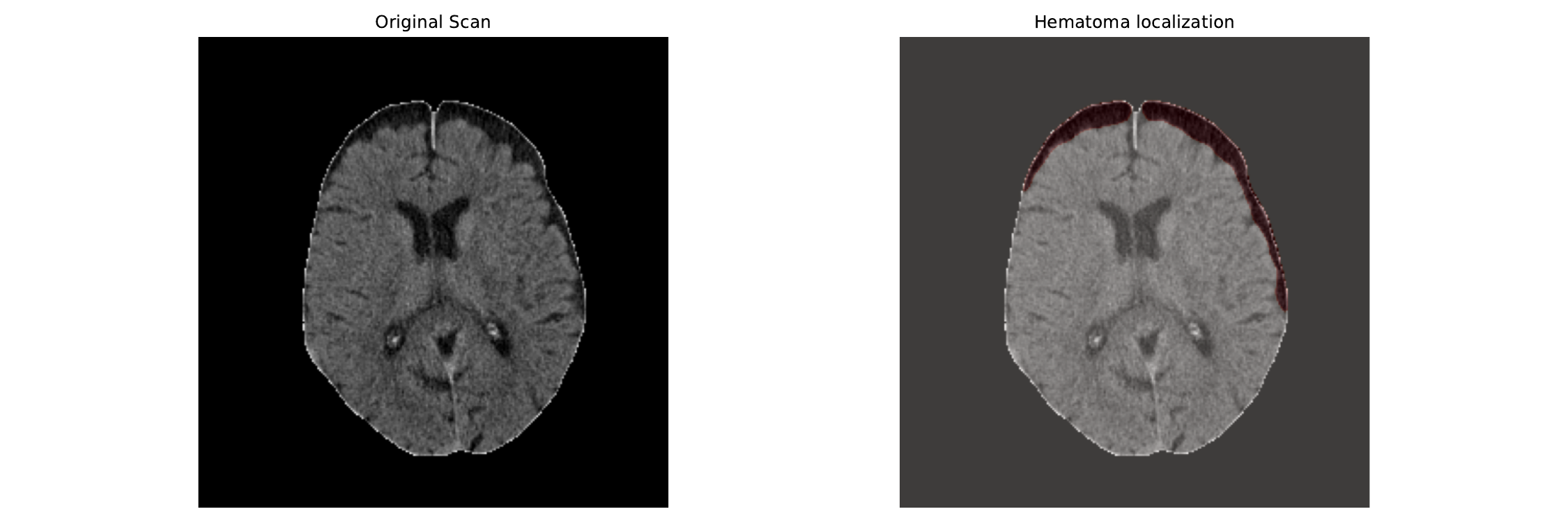}
        \label{fig:subfig1}
    \end{subfigure}


    \begin{subfigure}{\textwidth}
        \centering
        \includegraphics[width=0.9\textwidth]{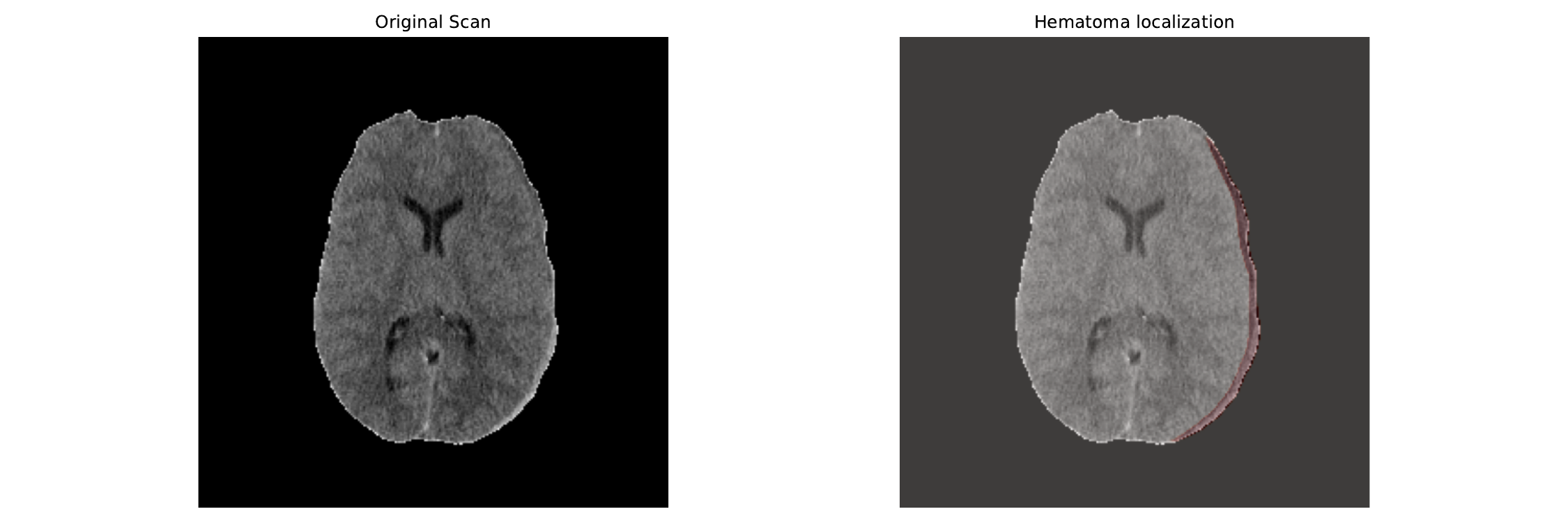}
        \label{fig:subfig2}
    \end{subfigure}


    \begin{subfigure}{\textwidth}
        \centering
        \includegraphics[width=0.9\textwidth]{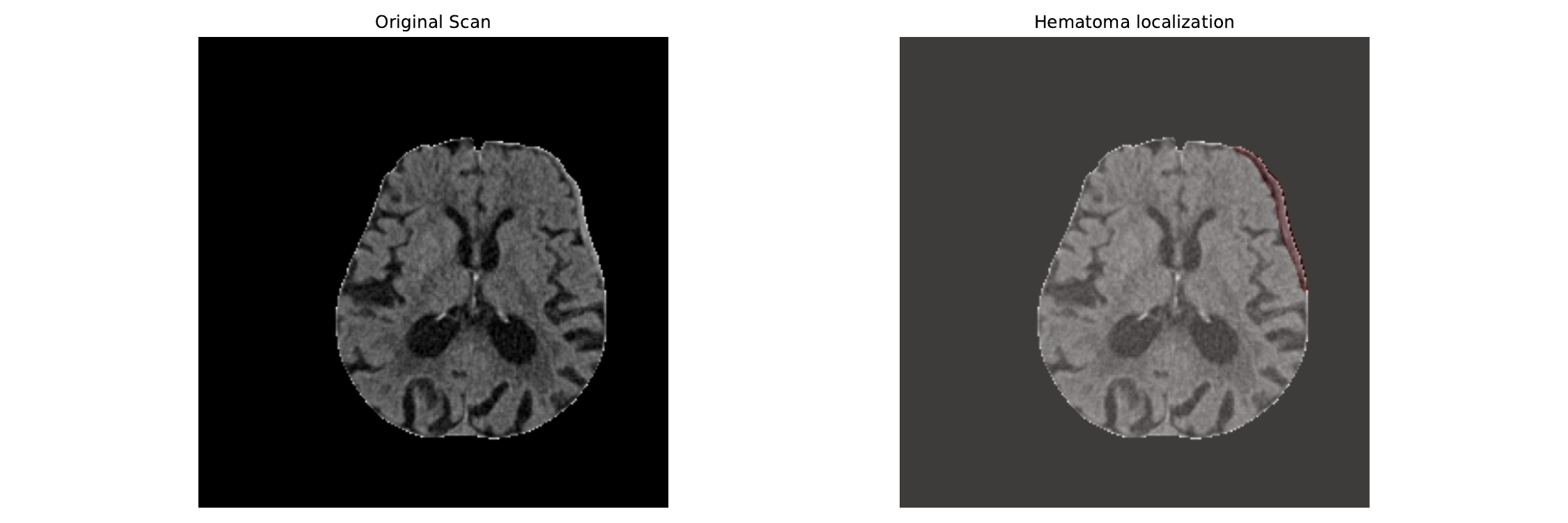}
        \label{fig:subfig3}
    \end{subfigure}


    \begin{subfigure}{\textwidth}
        \centering
        \includegraphics[width=0.9\textwidth]{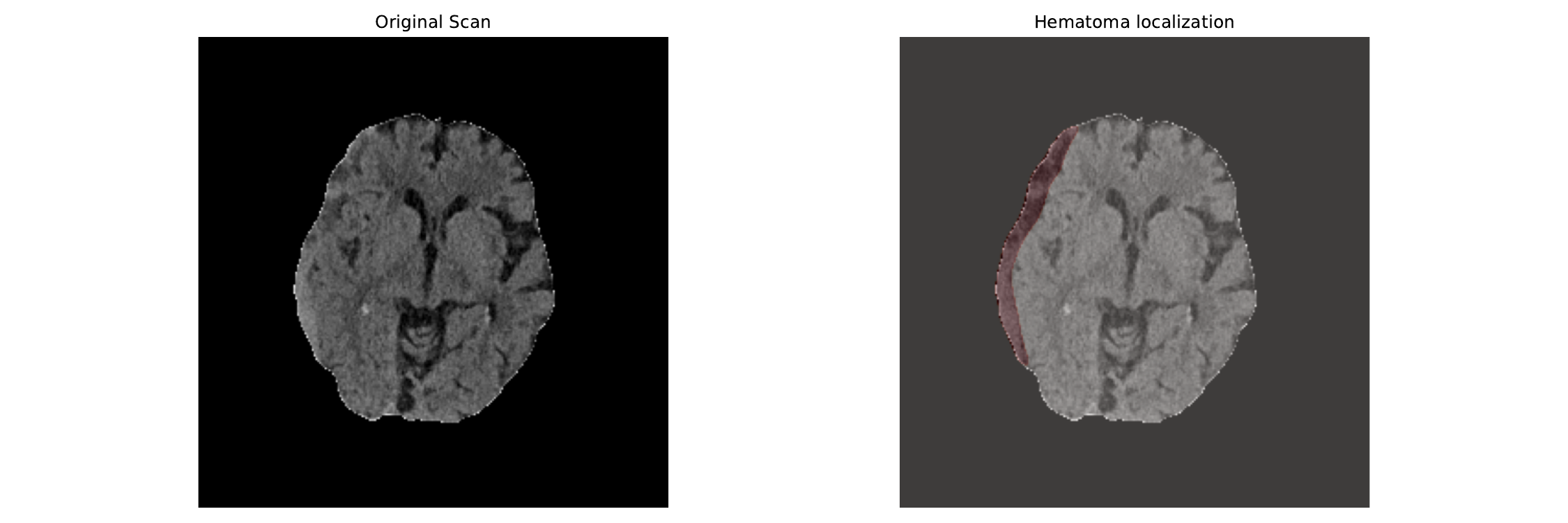}
        \label{fig:subfig4}
    \end{subfigure}

    \caption{Examples of subdural hematoma (SDH) localization test set results from the proposed segmentation model.}
    \label{fig:segmentation_example}
\end{figure*}

These results suggest that the hybrid design, combining Swin Transformer global attention with CNN-based local feature extraction, enables robust segmentation performance and generalization across imaging tasks. Representative examples of correctly segmented hematoma slices are shown in Figure \ref{fig:segmentation_example}, and out-of-sample quantitative comparisons with baseline models trained on the same training set as our proposed architecture are summarized in Table \ref{tab:segmentation_results}.

\begin{table}[H]
\centering
\caption{Segmentation model performance on the validation set. Results are reported as Dice scores.}
\label{tab:segmentation_results}
\small
\begin{tabular}{p{0.64\columnwidth}c}
\hline
\textbf{Model} & \textbf{SDH (Validation)} \\
\hline
Hybrid (proposed, multi-output fusion) & \textbf{0·768} \\
Vanilla Swin U-Net & 0·692 \\
Vanilla Swin U-Net w/ Frozen Layers &  0·720   \\
Attention U-Net w/ spatial \& channel attention & 0·741 \\
Hybrid (single-output fusion) & 0·754 \\
\hline
\end{tabular}
\end{table}

\subsection{Ensemble Selection Procedure} \label{subsec:ensemble}
\noindent
The ensemble was constructed using a greedy forward-selection procedure applied to the validation set. Rather than initializing the search from only the single best model, we began the greedy selection process from each of the top-performing individual models to reduce sensitivity to the initial one. For each initialization, models were iteratively added if their inclusion improved the ensemble’s mean validation AUC. After no further improvement was obtainable, the resulting ensembles were compared, and the best-performing subset was selected and subsequently evaluated on the independent test set. This approach ensures that performance gains arise from complementary model diversity rather than overfitting. Full per-model metrics and selection order are reported in \appModelSelection.


\section{Discussion}
\noindent
In this study, we developed a multimodal framework for detection and localization of subdural hematomas (SDH) that integrates tabular, imaging, and segmentation-based models into a unified predictive pipeline. Our approach combines a 3D convolutional neural network (CNN) for volumetric hematoma detection with a custom 2D hybrid segmentation model that merges convolutional and transformer-based components. This dual-path architecture enables both high-level classification across CT volumes and fine-grained localization within individual slices, bridging the gap between interpretability and diagnostic accuracy.

A major contribution of our work lies in integrating multiple specialized models (tabular, 3D CNN, and segmentation-based CNN) into a flexible ensemble that enhances robustness and performance. The ensemble design leverages complementary information from different data modalities, improving overall discrimination power compared with each single model. The custom segmentation network, which combines a U-Net–like encoder–decoder with Swin Transformer attention modules, produces anatomically precise probability maps, offering both clinical interpretability and high quantitative performance.

In contrast to existing commercial solutions, our framework offers full methodological transparency and comprehensive performance reporting across both detection and localization tasks. Unlike proprietary systems that restrict evaluation to hematomas above predefined size thresholds, our models are validated across the entire spectrum of hematoma presentations, highlighting the robustness and generalizability of our multimodal pipeline.

Our primary limitation is that the segmentation model was trained on a limited number of manually annotated scans, which may constrain generalizability. Expanding the pool of annotated images, particularly across diverse scanners and patient populations, will be essential for improving robustness. Leveraging semi-supervised or weakly supervised learning could further reduce dependence on manual labeling, making future iterations of the model more scalable. Additionally, external validation across institutions and imaging protocols will be critical for assessing real-world performance.

We plan to deploy this pipeline as a clinical decision support tool for prospective data analysis and real-time use. The framework is designed for seamless integration into electronic medical record (EMR) systems such as EPIC, providing automated, interpretable, and scalable SDH detection. By assisting clinicians in identifying and localizing subdural hematomas of any size without manual measurement constraints, our work moves toward practical, clinically deployable AI systems that enhance diagnostic accuracy and support timely, informed decision-making in neuroimaging workflows. With this system in place, we can begin to reduce the “time to surgery” penalty, improve consistency in management, and ultimately enhance patient outcomes in SDH care.





\section*{Contributors}
\noindent
VS, RK and BB had full access to all the 
data in the study and take responsibility for the integrity of the data and the 
accuracy of the data analysis. Concept and design: VS, RK, BB,  DO, TM, DB. Acquisition, analysis, or interpretation of data: 
VS, RK, BB. Drafting of the manuscript: VS, RK, BB. 
Critical revision of the manuscript for important intellectual content: All 
authors. Statistical analysis: VS, RK, BB. Administrative, 
technical, or clinical support: DO, TM. Supervision: TM, DB.

\section*{Declaration of interests}
\noindent
The authors declare no competing interests.

\section*{Data sharing}
\noindent
The data used in this study originate from the 
Hartford HealthCare system and are subject to institutional data use agreements and patient privacy protections. All analyses were performed within the Hartford HealthCare secure computing environment. Per institutional policy and regulatory requirements, these data cannot be transferred or shared outside the health system.

\section*{Acknowledgments}
\noindent
The authors thank the Hartford Hospital clinical teams for providing access to the patient data used in this study, and Hartford HealthCare for computational resources. The authors are also grateful to the clinicians (Diego Ojeda and Tapan Mehta) who offered domain feedback during model interpretation and validation and helped significantly with hematoma annotations. The authors would like to warmly thank Stephanie Tran at MIT for her invaluable support and continuous assistance in the administration of this project.

\section*{Declaration of generative AI and AI-assisted technologies in the manuscript preparation process}
\noindent
During the preparation of this work the author(s) used ChatGPT in order to fix syntax and grammar errors. After using this tool/service, the author(s) reviewed and edited the content as needed and take(s) full responsibility for the content of the published article.


\bibliography{references.bib} 

@article{kaya2023revolutionary,
  title={A revolutionary acute subdural hematoma detection based on two-tiered artificial intelligence model},
  author={Kaya, {\.I}smail and Gen{\c{c}}t{\"u}rk, Tu{\u{g}}rul Hakan and G{\"u}la{\u{g}}{\i}z, Fidan Kaya},
  journal={Turkish Journal of Trauma \& Emergency Surgery},
  volume={29},
  number={8},
  pages={858},
  year={2023},
  publisher={Turkish Association of Trauma and Emergency Surgery}
}

@article{colasurdo2022automated,
  title={Automated detection and analysis of subdural hematomas using a machine learning algorithm},
  author={Colasurdo, Marco and Leibushor, Nir and Robledo, Ariadna and Vasandani, Viren and Luna, Zean Aaron and Rao, Abhijit S and Garcia, Roberto and Srinivasan, Visish M and Sheth, Sunil A and Avni, Naama and others},
  journal={Journal of neurosurgery},
  volume={138},
  number={4},
  pages={1077--1084},
  year={2022},
  publisher={American Association of Neurological Surgeons}
}

@article{petrov2024ai,
  title={AI-Based Approach to One-Click Chronic Subdural Hematoma Segmentation Using Computed Tomography Images},
  author={Petrov, Andrey and Kashevnik, Alexey and Haleev, Mikhail and Ali, Ammar and Ivanov, Arkady and Samochernykh, Konstantin and Rozhchenko, Larisa and Bobinov, Vasiliy},
  journal={Sensors},
  volume={24},
  number={3},
  pages={721},
  year={2024},
  publisher={MDPI}
}

@article{Farzaneh2020,
author = {Farzaneh, Negar and Williamson, Craig and Jiang, Cheng and Srinivasan, Ashok and Bapuraj, Jayapalli and Gryak, Jonathan and Najarian, Kayvan and Soroushmehr, S.M.Reza},
year = {2020},
month = {09},
pages = {},
title = {Automated Segmentation and Severity Analysis of Subdural Hematoma for Patients with Traumatic Brain Injuries},
volume = {10},
journal = {Diagnostics (Basel, Switzerland)},
doi = {10.3390/diagnostics10100773}
}

@article{frontera2011national,
  title={National trend in prevalence, cost, and discharge disposition after subdural hematoma from 1998--2007},
  author={Frontera, Jennifer A and Egorova, Natalia and Moskowitz, Alan J},
  journal={Critical care medicine},
  volume={39},
  number={7},
  pages={1619--1625},
  year={2011},
  publisher={LWW}
}

@article{balser2015actual,
  title={Actual and projected incidence rates for chronic subdural hematomas in United States Veterans Administration and civilian populations},
  author={Balser, David and Farooq, Sameer and Mehmood, Talha and Reyes, Marleen and Samadani, Uzma},
  journal={Journal of neurosurgery},
  volume={123},
  number={5},
  pages={1209--1215},
  year={2015},
  publisher={American Association of Neurological Surgeons}
}

@article{rauhala2019incidence,
  title={The incidence of chronic subdural hematomas from 1990 to 2015 in a defined Finnish population},
  author={Rauhala, Minna and Luoto, Teemu M and Huhtala, Heini and Iverson, Grant L and Niskakangas, Tero and {\"O}hman, Juha and Hel{\'e}n, Pauli},
  journal={Journal of neurosurgery},
  volume={132},
  number={4},
  pages={1147--1157},
  year={2019},
  publisher={American Association of Neurological Surgeons}
}

@article{chen2024advances,
  title={Advances in chronic subdural hematoma and membrane imaging},
  author={Chen, Huanwen and Colasurdo, Marco and Malhotra, Ajay and Gandhi, Dheeraj and Bodanapally, Uttam K},
  journal={Frontiers in Neurology},
  volume={15},
  pages={1366238},
  year={2024},
  publisher={Frontiers Media SA}
}

@article{ryan2012acute,
  title={Acute traumatic subdural hematoma: current mortality and functional outcomes in adult patients at a Level I trauma center},
  author={Ryan, Christina G and Thompson, Rachel E and Temkin, Nancy R and Crane, Paul K and Ellenbogen, Richard G and Elmore, Joann G},
  journal={Journal of Trauma and Acute Care Surgery},
  volume={73},
  number={5},
  pages={1348--1354},
  year={2012},
  publisher={LWW}
}

@article{fluss2023deadliness,
  title={Deadliness of Traumatic Subdural Hematomas in the First Quarter of the Year: A Measurement by the American College of Surgeons-National Surgical Quality Improvement Program (ACS-NSQIP)},
  author={Fluss, Rose and Ryvlin, Jessica and Lam, Sharon and Abdullah, Muhammad and Altschul, David J},
  journal={Cureus},
  volume={15},
  number={12},
  year={2023},
  publisher={Cureus}
}

@article{salih2022reduced,
  title={Reduced recurrence of chronic subdural hematomas treated with open surgery followed by middle meningeal artery embolization compared to open surgery alone: a propensity score--matched analysis},
  author={Salih, Mira and Shutran, Max and Young, Michael and Vega, Rafael A and Stippler, Martina and Papavassiliou, Efstathios and Alterman, Ron L and Thomas, Ajith and Taussky, Philipp and Moore, Justin and others},
  journal={Journal of neurosurgery},
  volume={139},
  number={1},
  pages={124--130},
  year={2022},
  publisher={American Association of Neurological Surgeons}
}

@article{moshayedi2020middle,
  title={Middle meningeal artery embolization in chronic subdural hematoma: implications of pathophysiology in trial design},
  author={Moshayedi, Pouria and Liebeskind, David S},
  journal={Frontiers in Neurology},
  volume={11},
  pages={923},
  year={2020},
  publisher={Frontiers Media SA}
}

@article{colonna2025impact,
  title={Impact of surgical timing on chronic subdural hematoma outcomes: novel insights from a multicenter study},
  author={Colonna, Stefano and Lo Bue, Enrico and Pesaresi, Alessandro and Dolci, Lorenzo and Gatto, Andrea and Ceroni, Luca and Pesce, Alessandro and Salvati, Maurizio and Armocida, Daniele and Frati, Alessandro and others},
  journal={Neurosurgical Review},
  volume={48},
  number={1},
  pages={1--8},
  year={2025},
  publisher={Springer}
}

@article{chen2020impact,
  title={The impact of time from injury to surgery in functional recovery of traumatic acute subdural hematoma},
  author={Chen, Shih-Han and Sun, Jui-Ming and Fang, Wen-Kuei},
  journal={BMC neurology},
  volume={20},
  number={1},
  pages={226},
  year={2020},
  publisher={Springer}
}

@article{tien2011reducing,
  title={Reducing time-to-treatment decreases mortality of trauma patients with acute subdural hematoma},
  author={Tien, Homer CN and Jung, Vincent and Pinto, Ruxandra and Mainprize, Todd and Scales, Damon C and Rizoli, Sandro B},
  journal={Annals of surgery},
  volume={253},
  number={6},
  pages={1178--1183},
  year={2011},
  publisher={LWW}
}

@article{debs2024middle,
  title={Middle meningeal artery embolization following surgical evacuation of symptomatic chronic subdural hematoma improves outcomes, interim results of a prospective randomized trial},
  author={Debs, Luca H and Vale, Fernando L and Walker, Samantha and Toro, Diana and Mansouri, Seena and Macomson, Samuel D and Rahimi, Scott Y},
  journal={Journal of Clinical Neuroscience},
  volume={128},
  pages={110783},
  year={2024},
  publisher={Elsevier}
}

@article{davies2024adjunctive,
  title={Adjunctive middle meningeal artery embolization for subdural hematoma},
  author={Davies, Jason M and Knopman, Jared and Mokin, Maxim and Hassan, Ameer E and Harbaugh, Robert E and Khalessi, Alexander and Fiehler, Jens and Gross, Bradley A and Grandhi, Ramesh and Tarpley, Jason and others},
  journal={New England Journal of Medicine},
  volume={391},
  number={20},
  pages={1890--1900},
  year={2024},
  publisher={Mass Medical Soc}
}

@article{ramachandran2007chronic,
  title={Chronic subdural hematomas—causes of morbidity and mortality},
  author={Ramachandran, Ramnarayan and Hegde, Thimmappa},
  journal={Surgical neurology},
  volume={67},
  number={4},
  pages={367--372},
  year={2007},
  publisher={Elsevier}
}

@article{miah2021radiological,
  title={Radiological prognostic factors of chronic subdural hematoma recurrence: a systematic review and meta-analysis},
  author={Miah, Ishita P and Tank, Yeliz and Rosendaal, Frits R and Peul, Wilco C and Dammers, Ruben and Lingsma, Hester F and den Hertog, Heleen M and Jellema, Korn{\'e} and van der Gaag, Niels A and Dutch Chronic Subdural Hematoma Research Group},
  journal={Neuroradiology},
  volume={63},
  number={1},
  pages={27--40},
  year={2021},
  publisher={Springer}
}

@article{oh2010postoperative,
  title={Postoperative course and recurrence of chronic subdural hematoma},
  author={Oh, Hyuck-Jin and Lee, Kyeong-Seok and Shim, Jae-Jun and Yoon, Seok-Mann and Yun, Il-Gyu and Bae, Hack-Gun},
  journal={Journal of Korean Neurosurgical Society},
  volume={48},
  number={6},
  pages={518},
  year={2010}
}

@article{pinto2023artificial,
  title={How artificial intelligence is shaping medical imaging technology: a survey of innovations and applications},
  author={Pinto-Coelho, Lu{\'\i}s},
  journal={Bioengineering},
  volume={10},
  number={12},
  pages={1435},
  year={2023},
  publisher={MDPI}
}

@article{khalifa2024ai,
  title={AI in diagnostic imaging: revolutionising accuracy and efficiency},
  author={Khalifa, Mohamed and Albadawy, Mona},
  journal={Computer Methods and programs in biomedicine update},
  volume={5},
  pages={100146},
  year={2024},
  publisher={Elsevier}
}

@article{seyam2022utilization,
  title={Utilization of artificial intelligence--based intracranial hemorrhage detection on emergent noncontrast CT images in clinical workflow},
  author={Seyam, Muhannad and Weikert, Thomas and Sauter, Alexander and Brehm, Alex and Psychogios, Marios-Nikos and Blackham, Kristine A},
  journal={Radiology: Artificial Intelligence},
  volume={4},
  number={2},
  pages={e210168},
  year={2022},
  publisher={Radiological Society of North America}
}

@article{abed2025artificial,
  title={Artificial intelligence for detecting traumatic intracranial haemorrhage with CT: A workflow-oriented implementation},
  author={Abed, Selim and Hergan, Klaus and Pfaff, Johannes and D{\"o}rrenberg, Jan and Brandstetter, Lucas and Gradl, Johann},
  journal={The Neuroradiology Journal},
  pages={19714009251346477},
  year={2025},
  publisher={SAGE Publications Sage UK: London, England}
}

@article{rohren2025post,
  title={Post-deployment Monitoring of AI Performance in Intracranial Hemorrhage Detection by ChatGPT},
  author={Rohren, Eric and Ahmadzade, Mohadese and Colella, Sofia and Kottler, Nina and Krishnan, Sriyesh and Poff, Jason and Rastogi, Neelesh and Wiggins, Walter and Yee, Joyce and Zuluaga, Carlos and others},
  journal={Academic Radiology},
  year={2025},
  publisher={Elsevier}
}

@article{bayar2025artificial,
  title={Artificial intelligence in radiology: diagnostic sensitivity of ChatGPT for detecting hemorrhages in cranial computed tomography scans},
  author={Bayar-Kap{\i}c{\i}, Olga and Altun{\i}{\c{s}}{\i}k, Erman and Musabeyo{\u{g}}lu, Feyza and Dev, {\c{S}}eyda and Kaya, {\"O}mer},
  journal={Diagnostic and Interventional Radiology},
  year={2025},
  publisher={Diagnostic and Interventional Radiology}
}

@article{zarei2024deep,
  title={Do Deep Learning Algorithms Accurately Segment Intracerebral Hemorrhages on Noncontrast Computed Tomography? A Systematic Review and Meta-Analysis},
  author={Zarei, Diana and Issaiy, Mahbod and Kolahi, Shahriar and Liebeskind, David S},
  journal={Stroke: Vascular and Interventional Neurology},
  volume={4},
  number={4},
  pages={e001314},
  year={2024}
}

@article{hirata2025brain,
  title={Brain Hematoma Marker Recognition Using Multitask Learning: SwinTransformer and Swin-Unet},
  author={Hirata, Kodai and Okita, Tsuyoshi},
  journal={arXiv preprint arXiv:2505.06185},
  year={2025}
}

@article{soenksen2022integrated,
  title={Integrated multimodal artificial intelligence framework for healthcare applications},
  author={Soenksen, Luis R and Ma, Yu and Zeng, Cynthia and Boussioux, Leonard and Villalobos Carballo, Kimberly and Na, Liangyuan and Wiberg, Holly M and Li, Michael L and Fuentes, Ignacio and Bertsimas, Dimitris},
  journal={NPJ digital medicine},
  volume={5},
  number={1},
  pages={149},
  year={2022},
  publisher={Nature Publishing Group UK London}
}

@incollection{LEV2002427,
title = {17 - CT Angiography and CT Perfusion Imaging},
editor = {Arthur W. Toga and John C. Mazziotta},
booktitle = {Brain Mapping: The Methods (Second Edition)},
publisher = {Academic Press},
edition = {Second Edition},
address = {San Diego},
pages = {427-484},
year = {2002},
isbn = {978-0-12-693019-1},
doi = {https://doi.org/10.1016/B978-012693019-1/50019-8},
url = {https://www.sciencedirect.com/science/article/pii/B9780126930191500198},
author = {M.H. Lev and R.G. Gonzalez}
}

@inproceedings{chen2016xgboost,
  title={Xgboost: A scalable tree boosting system},
  author={Chen, Tianqi and Guestrin, Carlos},
  booktitle={Proceedings of the 22nd acm sigkdd international conference on knowledge discovery and data mining},
  pages={785--794},
  year={2016}
}

@article{charlson1987new,
  title={A new method of classifying prognostic comorbidity in longitudinal studies: development and validation},
  author={Charlson, Mary E and Pompei, Peter and Ales, Kathy L and MacKenzie, C Ronald},
  journal={Journal of chronic diseases},
  volume={40},
  number={5},
  pages={373--383},
  year={1987},
  publisher={Elsevier}
}

@article{quan2005coding,
  title={Coding algorithms for defining comorbidities in ICD-9-CM and ICD-10 administrative data},
  author={Quan, Hude and Sundararajan, Vijaya and Halfon, Patricia and Fong, Andrew and Burnand, Bernard and Luthi, Jean-Christophe and Saunders, L Duncan and Beck, Cynthia A and Feasby, Thomas E and Ghali, William A},
  journal={Medical care},
  volume={43},
  number={11},
  pages={1130--1139},
  year={2005},
  publisher={LWW}
}

@article{loshchilov2017decoupled,
  title={Decoupled weight decay regularization},
  author={Loshchilov, Ilya and Hutter, Frank},
  journal={arXiv preprint arXiv:1711.05101},
  year={2017}
}

@misc{buda2019lggmri,
  author       = {Mateusz Buda},
  title        = {LGG MRI Segmentation Dataset},
  howpublished = {\url{https://www.kaggle.com/datasets/mateuszbuda/lgg-mri-segmentation}},
  year         = {2018}
}




\end{document}


\begin{appendices}
\renewcommand{\thesection}{\arabic{section}}


\mainheading{Supplementary Material}

\tableofcontents
\newpage

\section{Data}

\subsection{Cohort statistics}

\begin{table}[H]
\centering
\scriptsize
\caption{Baseline characteristics of the study cohort stratified by presence of subdural hematoma (SDH). Continuous variables are summarized as mean $\pm$ SD and compared using Welch’s $t$-tests; categorical variables are compared using $\chi^2$ tests. Missing values are reported as percentages.}
\resizebox{\textwidth}{!}{
\begin{tabular}{lcccc}
\hline
\textbf{Variable} & \textbf{No SDH} & \textbf{SDH} & \textbf{Missing (\%)} & \textbf{$p$-value} \\
\hline
\multicolumn{5}{l}{\textbf{Platelet and Coagulation Studies}} \\
Platelet count (most recent) & 232·71 $\pm$ 91·16 & 220·73 $\pm$ 100·97 & 23·1 & $<$0·001 \\
Platelet count (weighted) & 230·15 $\pm$ 83·52 & 215·40 $\pm$ 90·43 & 22·9 & $<$0·001 \\
INR (most recent) & 1·30 $\pm$ 0·66 & 1·24 $\pm$ 0·59 & 50·9 & $<$0·001 \\
INR (weighted) & 1·31 $\pm$ 0·61 & 1·28 $\pm$ 0·54 & 50·0 & $<$0·001 \\
PT (most recent) & 14·83 $\pm$ 7·48 & 14·15 $\pm$ 6·39 & 50·0 & 0·004 \\
PT (weighted) & 14·96 $\pm$ 6·97 & 14·64 $\pm$ 6·08 & 50·5 & 0·024 \\
aPTT (most recent) & 35·48 $\pm$ 22·79 & 32·13 $\pm$ 19·36 & 74·5 & $<$0·001 \\
aPTT (weighted) & 35·35 $\pm$ 20·30 & 32·43 $\pm$ 10·94 & 74·4 & $<$0·001 \\
Anti-Xa (most recent) & 0·46 $\pm$ 0·39 & 0·41 $\pm$ 0·42 & 92·0 & 0·023 \\
Anti-Xa (weighted) & 0·48 $\pm$ 0·32 & 0·44 $\pm$ 0·38 & 92·0 & 0·023 \\
\hline
\multicolumn{5}{l}{\textbf{Anticoagulant and Antiplatelet Medications}} \\
Heparin use & 0·10 $\pm$ 0·30 & 0·12 $\pm$ 0·33 & 0·0 & $<$0·001 \\
Aspirin use & 0·23 $\pm$ 0·47 & 0·30 $\pm$ 0·46 & 0·0 & $<$0·001 \\
Apixaban use & 0·06 $\pm$ 0·28 & 0·07 $\pm$ 0·27 & 0·0 & $<$0·001 \\
Clopidogrel use & 0·07 $\pm$ 0·25 & 0·05 $\pm$ 0·22 & 0·0 & $<$0·001 \\
Warfarin use & 0·05 $\pm$ 0·22 & 0·07 $\pm$ 0·25 & 0·0 & $<$0·001 \\
Ticagrelor use & 0·01 $\pm$ 0·09 & 0·01 $\pm$ 0·10 & 0·0 & 0·211 \\
Dabigatran use & 0·00 $\pm$ 0·07 & 0·01 $\pm$ 0·08 & 0·0 & 0·224 \\
Rivaroxaban use & 0·01 $\pm$ 0·12 & 0·01 $\pm$ 0·12 & 0·0 & 0·028 \\
\hline
\multicolumn{5}{l}{\textbf{Charlson Comorbidity Index and Clinical Risk}} \\
Charlson index (days since computation) & 299·46 $\pm$ 535·31 & 243·22 $\pm$ 541·14 & 29·3 & $<$0·001 \\
Age (score according to ACCI) & 2·17 $\pm$ 1·52 & 2·72 $\pm$ 1·31 & 0·0 & $<$0·001 \\
\hline
\multicolumn{5}{l}{\textbf{Charlson Comorbidity Index Components}} \\
Acute myocardial infarction (AMI) & 0·01 $\pm$ 0·10 & 0·01 $\pm$ 0·12 & 0·0 & 0·094 \\
Congestive heart failure (CHF) & 0·04 $\pm$ 0·19 & 0·04 $\pm$ 0·21 & 0·0 & 0·154 \\
Peripheral vascular disease (PVD) & 0·04 $\pm$ 0·20 & 0·03 $\pm$ 0·17 & 0·0 & $<$0·001 \\
Cerebrovascular disease (CEVD) & 0·07 $\pm$ 0·26 & 0·08 $\pm$ 0·27 & 0·0 & 0·110 \\
Dementia & 0·03 $\pm$ 0·17 & 0·02 $\pm$ 0·15 & 0·0 & 0·026 \\
Chronic obstructive pulmonary disease (COPD) & 0·03 $\pm$ 0·18 & 0·03 $\pm$ 0·17 & 0·0 & $<$0·001 \\
Rheumatologic disease & 0·01 $\pm$ 0·11 & 0·02 $\pm$ 0·15 & 0·0 & $<$0·001 \\
Peptic ulcer disease & 0·01 $\pm$ 0·10 & 0·01 $\pm$ 0·08 & 0·0 & 0·594 \\
Mild liver disease & 0·01 $\pm$ 0·11 & 0·02 $\pm$ 0·15 & 0·0 & 0·050 \\
Moderate–severe liver disease & 0·02 $\pm$ 0·23 & 0·02 $\pm$ 0·22 & 0·0 & 0·594 \\
Diabetes without complication & 0·09 $\pm$ 0·41 & 0·08 $\pm$ 0·39 & 0·0 & 0·139 \\
Diabetes with complication & 0·04 $\pm$ 0·19 & 0·04 $\pm$ 0·23 & 0·0 & 0·338 \\
Hemiplegia/paraplegia & 0·02 $\pm$ 0·21 & 0·02 $\pm$ 0·23 & 0·0 & 0·336 \\
Renal disease & 0·05 $\pm$ 0·31 & 0·05 $\pm$ 0·33 & 0·0 & 0·348 \\
Metastatic cancer & 0·09 $\pm$ 0·73 & 0·09 $\pm$ 0·71 & 0·0 & 0·744 \\
Any malignancy (non-metastatic) & 0·21 $\pm$ 0·61 & 0·20 $\pm$ 0·60 & 0·0 & 0·611 \\
AIDS/HIV & 0·02 $\pm$ 0·36 & 0·02 $\pm$ 0·34 & 0·0 & 0·596 \\
\hline
\multicolumn{5}{l}{\textbf{Demographics}} \\
Sex (0 = Male, 1 = Female) & 0·52 $\pm$ 0·50 & 0·41 $\pm$ 0·49 & 0·0 & $<$0·001 \\
\hline
\end{tabular}}
\label{tab:cohort_stats}
\end{table}


\subsection{Charlson Comorbidity Index (CCI)}
\noindent
The Charlson Comorbidity Index (CCI) was incorporated to quantify each patient’s comorbidity burden using a weighted aggregation of clinically significant chronic conditions. Each condition, such as myocardial infarction, congestive heart failure, diabetes, renal disease, and malignancy, is assigned an importance weight that reflects its empirically estimated association with long-term mortality, as established in the original Charlson cohort~\cite{charlson1987new} and subsequent validation studies~\cite{quan2005coding}. These weights capture the relative prognostic impact of each comorbidity on survival risk, ensuring that conditions with stronger mortality implications contribute more heavily to the overall score. 

To account for age-related risk, we adopted the Age-Adjusted Charlson Comorbidity Index (ACCI), in which additional points are added for older patients according to the following rule: one additional point for each decade of life over 40 years (i.e., +1 for ages 50–59, +2 for 60–69, +3 for 70–79, +4 for 80–89, and so on). This modification integrates the well-established impact of age on mortality risk, providing a more accurate measure of baseline health status. 

We retained the original Charlson weights for each comorbidity, as they provide a validated, interpretable measure that has been extensively adopted in clinical research and risk-adjusted modeling. This approach allows our models to contextualize patient frailty and comorbidity severity.

\begin{table}[H]
\centering
\caption{Charlson Comorbidity Index (CCI) components and assigned weights, including age adjustment.}
\label{tab:charlson_index}
\begin{tabular}{llc}
\toprule
\textbf{Category} & \textbf{Condition} & \textbf{Weight} \\
\midrule
\textbf{Cardiovascular} & Myocardial infarction & 1 \\ 
& Congestive heart failure & 1 \\ 
& Peripheral vascular disease & 1 \\ 
& Cerebrovascular disease (stroke/TIA) & 1 \\ 
\textbf{Respiratory} & Chronic pulmonary disease (COPD) & 1 \\ 
\textbf{Endocrine} & Diabetes mellitus (without complications) & 1 \\ 
& Diabetes with chronic complications & 2 \\ 
\textbf{Renal/Hepatic} & Mild liver disease & 1 \\ 
& Moderate or severe liver disease & 3 \\ 
& Renal disease & 2 \\ 
\textbf{Oncologic} & Any malignancy (excluding metastasis) & 2 \\ 
& Metastatic solid tumor & 6 \\ 
& Leukemia or lymphoma & 2 \\ 
\textbf{Neurologic} & Dementia & 1 \\ 
& Hemiplegia or paraplegia & 2 \\ 
\textbf{Immunologic} & AIDS/HIV & 6 \\ 
\textbf{Other} & Connective tissue disease / rheumatologic disorder & 1 \\ 
\textbf{Age Adjustment} & Age 50–59 years & +1 \\ 
& Age 60–69 years & +2 \\ 
& Age 70–79 years & +3 \\ 
& Age 80–89 years & +4 \\ 
& Age $\geq$90 years & +5 \\
\bottomrule
\end{tabular}
\end{table}

In our analysis, rather than using the summed CCI score directly, each comorbidity component and the age adjustment were entered as separate features using their respective Charlson weights. This design preserves interpretability while allowing the model to learn non-linear interactions between age, specific comorbidities, and the remaining clinical features.


\section{3D CNN Architecture for Subdural Hematoma Detection}
\label{app:cnn_architecture}
\noindent
The architecture consists of five convolutional blocks, each containing a 3D convolutional layer, batch normalization, and a pooling operation. Blocks 3 and 4 incorporate skip connections from Blocks 1 and 2, respectively, which enhance gradient flow and promote multi-scale feature reuse. The use of $3\times3\times3$ kernels enables fine-grained feature extraction from volumetric data while maintaining computational efficiency. Since the model processes 3D inputs, which substantially increase the number of trainable parameters, the original CT volumes ($15\times512\times512$) were downsampled to $15\times256\times256$ to balance spatial resolution and computational scalability. 

\textbf{Convolutional Layers:} Five sequential 3D convolutional blocks with batch normalization and ReLU activation. The use of small $3\times3\times3$ kernels with padding preserves spatial dimensions. 

\textbf{Pooling:} Max pooling progressively reduces spatial resolution. Early layers pool only along the height and width dimensions to preserve depth information; deeper layers pool across all dimensions to capture full 3D spatial hierarchies. 

\textbf{Batch Normalization:} Applied after each convolution to stabilize training, accelerate convergence, and improve generalization. 

\textbf{Skip Connections:} Feature maps from Blocks 1 and 2 are concatenated with those from Blocks 3 and 4, respectively (after trilinear interpolation for alignment). These connections preserve low-level detail and improve gradient propagation through the network. 

\textbf{Dropout:} Applied at 10–25\% in selected layers to mitigate overfitting. 

\textbf{Fully Connected Layers:} Three dense layers (512–64–1) progressively reduce feature dimensionality and output the final binary logit for SDH classification. 

\begin{figure}[htbp]
\centering
\includegraphics[width=0.99\textwidth]{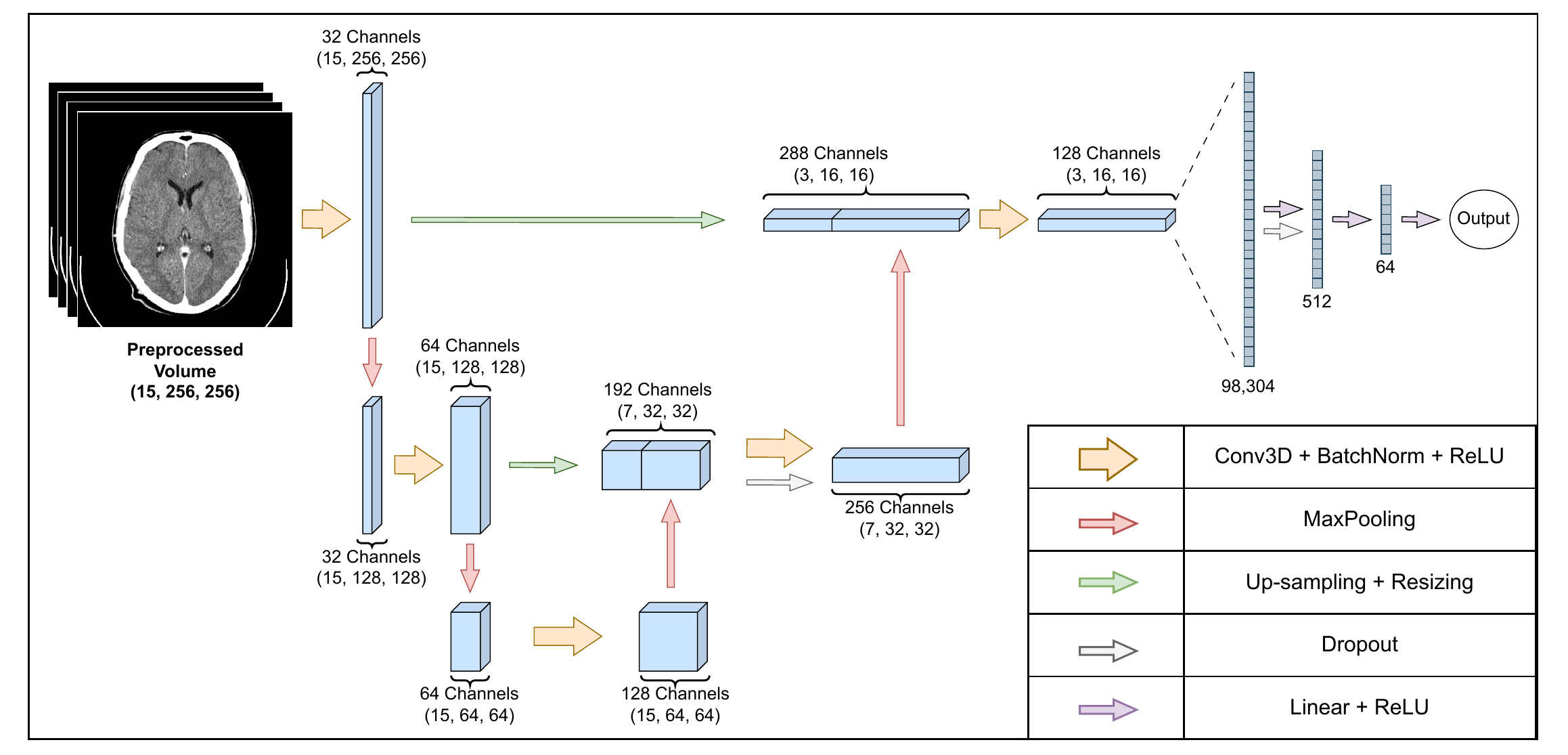}
\caption{Schematic of the custom 3D convolutional neural network for subdural hematoma classification.}
\label{fig:cnn_architecture}
\end{figure}

A schematic of the complete architecture is provided in \figurename ~\ref{fig:cnn_architecture}. The model was trained for a few epochs using a batch size of 16 and the AdamW optimizer \cite{loshchilov2017decoupled} (learning rate $5\times10^{-5}$). The checkpoint achieving the best validation AUC was selected for final evaluation.

\section{Segmentation Architecture}
\label{app:segmentation_architecture}
\noindent
Accurate segmentation of subdural hematomas requires capturing both local and global context. Local pixel-level features, such as sharp intensity changes near the cortical surface, may indicate hemorrhage, while global patterns like midline shift or hemispheric asymmetry reflect mass effect. To model both, we designed a hybrid architecture that combines a convolutional encoder–decoder backbone with a Swin Transformer module, leveraging convolutional layers for local feature extraction and self-attention for long-range contextual reasoning. 

The convolutional encoder consists of four residual blocks with progressively increasing feature dimensions (32, 64, 128, 256), each followed by max pooling for spatial downsampling. Each residual block includes three convolutional layers with batch normalization, LeakyReLU activation, and dropout regularization, facilitating deep feature learning while maintaining gradient stability. A bottleneck block with 512 channels serves as the deepest representation. 

The decoder mirrors the encoder with four up-convolutional blocks for spatial upsampling. Each upsampling stage incorporates an attention-based skip connection that fuses encoder and decoder features. The attention blocks perform both channel and spatial recalibration using global average and max pooling, emphasizing informative regions while suppressing background noise. Final decoder layers employ $1 \times 1$ convolutions to generate intermediate feature maps and the primary segmentation output. 

To enhance representational capacity beyond local convolutional receptive fields, a parallel Swin Transformer branch operates on non-overlapping image patches. The transformer encoder applies shifted-window self-attention, multi-head feature aggregation, and patch-merging downsampling to hierarchically encode global structure. The decoder reverses this hierarchy using patch-expansion layers, reconstructing spatial detail through learned upsampling and concatenation with earlier feature maps. Linear reduction layers convert transformer feature maps into single-channel representations aligned with the convolutional path. 

\begin{figure}[htbp]
\centering
\includegraphics[width=0.99\textwidth]{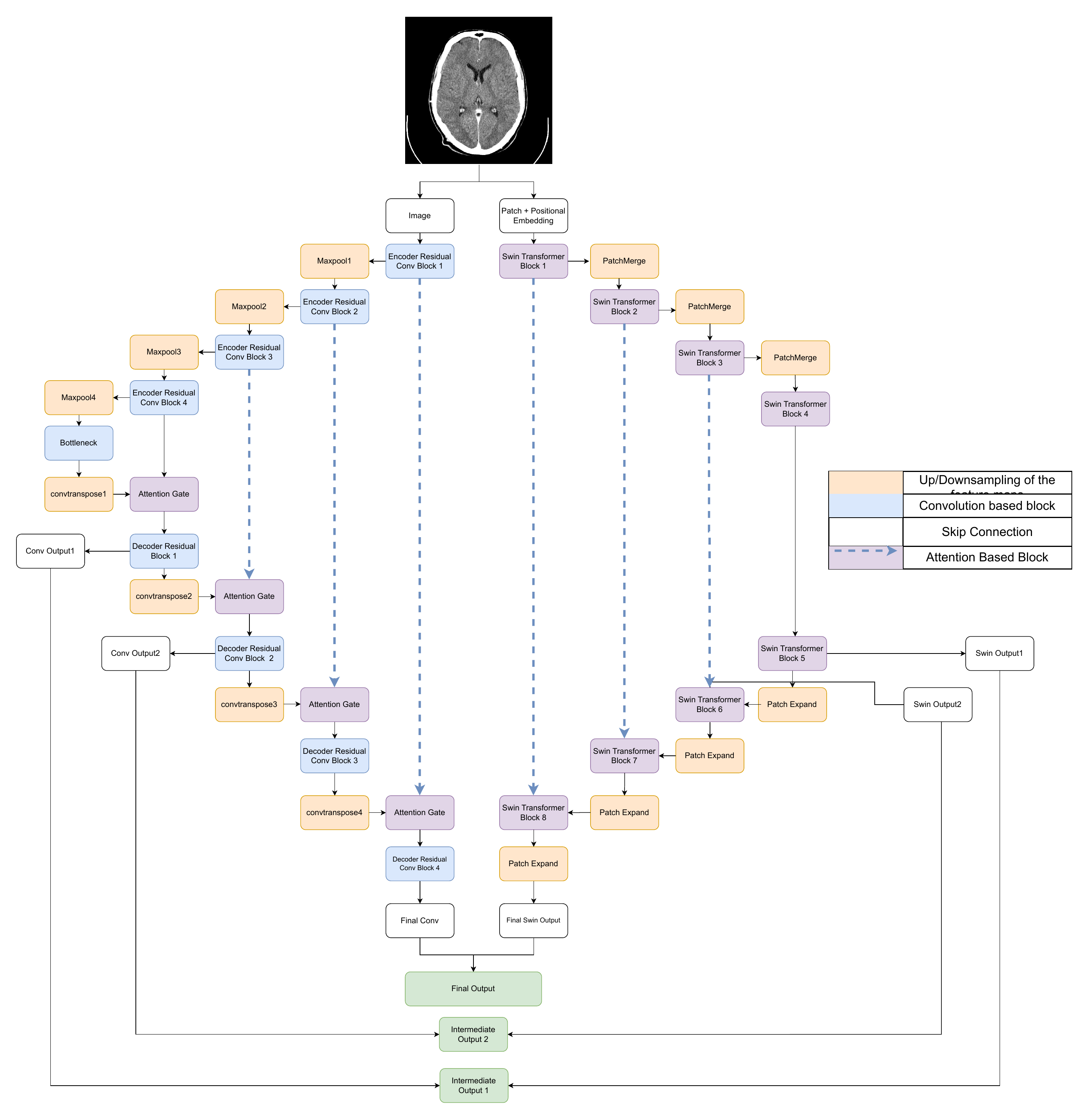}
\caption{Schematic of the custom segmentation model for subdural hematoma localization.}
\label{fig:seg_architecture}
\end{figure}

Outputs from both the convolutional decoder and Swin Transformer decoder are merged via $1 \times 1$ convolutions at multiple scales and fused into a final prediction map using an additional $1 \times 1$ convolution. The network outputs three segmentation maps (two intermediate and one final) representing hematoma probability at different feature depths. A schematic of the complete architecture is provided in \figurename~\ref{fig:seg_architecture}.

The model was trained using a composite loss function consisting of Dice loss and binary cross-entropy, with $\lambda =$ 0·6 to balance region overlap and voxel-wise accuracy: 

\begin{equation} 
\mathcal{L} = \lambda \mathcal{L}_{\text{Dice}} + (1 - \lambda) \, \mathcal{L}_{\text{BCE}}. 
\end{equation} 

Here, the Dice loss measures the overlap between estimated and ground-truth masks and is defined as: 

\begin{equation} 
\mathcal{L}_{\text{Dice}} = 1 - \frac{2 \sum_{i} p_i g_i}{\sum_{i} p_i^2 + \sum_{i} g_i^2 + \epsilon},
\end{equation} 

where $p_i$ denotes the predicted probability for pixel $i$, $g_i$ the ground-truth label for pixel $i$ (binary: 0 or 1) and $\epsilon$ a small constant added to prevent division by zero. 

Input CT slices were resampled to $256 \times 256$ resolution with three channels. Data augmentation included random flips and rotations. The model was optimized with AdamW \cite{loshchilov2017decoupled} using an initial learning rate of $10^{-4}$ and a batch size of 16, and trained for 100 epochs with early stopping based on validation Dice score.


\section{Model Selection and Ensembling Procedure}
\label{app:model_selection}
\noindent
To construct the final ensemble, we used a greedy forward-selection strategy based on validation AUC. We first evaluated all individual models and identified the top-performing candidates. To reduce sensitivity to the choice of starting model, the greedy procedure was initialized separately from each of the top five single models. For each initialization, models were iteratively added if their inclusion improved the ensemble’s mean prediction AUC on the validation set. The search terminated when no remaining model improved AUC. The best-performing ensemble across all initializations was then evaluated on the held-out test set.

This strategy was applied across all available model outputs, including the tabular XGBoost model, five 3D CNNs trained directly on CT volumes, and four CNNs trained on segmentation-derived masks. The optimal ensemble identified during validation was then evaluated on the held-out test set, yielding the final performance metrics reported in \tabClassificationResults. This approach balances predictive gain with simplicity, reducing the risk of overfitting while leveraging the diversity of model architectures and modalities.

\begin{algorithm}[H]
\caption{Greedy Ensemble Selection With Multiple Starting Seeds}
\label{alg:ensemble_selection}
\begin{algorithmic}[1]

\State Let $\mathcal{M}$ be the set of all candidate models.
\State Compute single-model AUCs on the validation set.
\State Let $\mathcal{S}$ be the set of the top-$k$ models by single-model AUC (e.g., $k=5$).

\State Initialize: best\_ensemble $\leftarrow \emptyset$, best\_AUC $\leftarrow 0$.

\For{each seed model $s \in \mathcal{S}$}
    \State $S \leftarrow \{s\}$  
    \State current\_AUC $\leftarrow$ AUC$(s)$
    \Repeat
        \State improvement $\leftarrow$ False
        \For{each model $m \in \mathcal{M} \setminus S$}
            \State Compute validation AUC of ensemble formed by averaging predictions of $S \cup \{m\}$
            \If{AUC improves over current\_AUC}
                \State Record best\_candidate $\leftarrow m$
                \State Record best\_candidate\_AUC
            \EndIf
        \EndFor
        \If{an improving candidate exists}
            \State $S \leftarrow S \cup \{ \text{best\_candidate} \}$
            \State current\_AUC $\leftarrow$ best\_candidate\_AUC
            \State improvement $\leftarrow$ True
        \EndIf
    \Until{improvement = False}

    \If{current\_AUC > best\_AUC}
        \State best\_ensemble $\leftarrow S$
        \State best\_AUC $\leftarrow$ current\_AUC
    \EndIf
\EndFor

\State \textbf{Output:} best\_ensemble and its test-set AUC.

\end{algorithmic}
\end{algorithm}

\bibliographystyle{ama} 
\bibliography{references.bib} 
\end{appendices}